\documentclass[10pt,twocolumn,letterpaper]{article}

\usepackage{cvpr}
\usepackage{times}
\usepackage{epsfig}
\usepackage{graphicx}
\usepackage{amsmath}
\usepackage{amssymb}
\usepackage{color}
\usepackage{bbding}
\usepackage{pifont}
\usepackage{xcolor}
\usepackage{colortbl}
\usepackage{mathtools}
\usepackage{mathrsfs}
\usepackage{multirow}
\usepackage{algorithmic}
\usepackage{algorithm}
\usepackage{array}
\usepackage{tabu}
\usepackage{adjustbox}
\usepackage{tabularx}
\usepackage{booktabs}
\usepackage{diagbox}
\usepackage[breaklinks=true,bookmarks=false]{hyperref}

\cvprfinalcopy 

\def\cvprPaperID{5493} 

\ifcvprfinal\fi
\begin{document}

\title{Pose-guided Visible Part Matching for Occluded Person ReID}


\author{Shang Gao\textsuperscript{1}, Jingya Wang\textsuperscript{3}, Huchuan Lu\textsuperscript{1,2\thanks{Corresponding Author}},  Zimo Liu\textsuperscript{1}\\
${}^1$Dalian University of Technology, 
${}^2$Pengcheng Lab\\
${}^3$UBTECH Sydney AI Center, The University of Sydney\\
{\tt\small gs940601k@gmail.com, jingya.wang@sydney.edu.au, lhchuan@dlut.edu.cn, lzm920316@gmail.com}
}

\maketitle

\begin{abstract}
   Occluded person re-identification is a challenging task as the appearance varies substantially with various obstacles, especially in the crowd scenario.
   To address this issue, we propose a Pose-guided Visible Part Matching (PVPM) method that jointly learns the discriminative features with pose-guided attention and self-mines the part visibility in an end-to-end framework.
   Specifically, the proposed PVPM includes two key components: 1) pose-guided attention (PGA) method for part
   feature pooling that exploits more discriminative local features; 2) pose-guided visibility predictor (PVP) that estimates whether a part suffers the occlusion or not.
   %
   %
   As there are no ground truth training annotations for the occluded part, we turn to utilize the characteristic of part correspondence in positive pairs and self-mining the correspondence scores via graph matching. The generated correspondence scores are then utilized as pseudo-labels for visibility predictor (PVP).  
   Experimental results on three reported occluded benchmarks show that the proposed method achieves competitive performance to state-of-the-art methods. The source codes are available at \url{https://github.com/hh23333/PVPM}
       \end{abstract}
       \section{Introduction}

\begin{figure}
          \centering
          \includegraphics[width=0.98\linewidth]{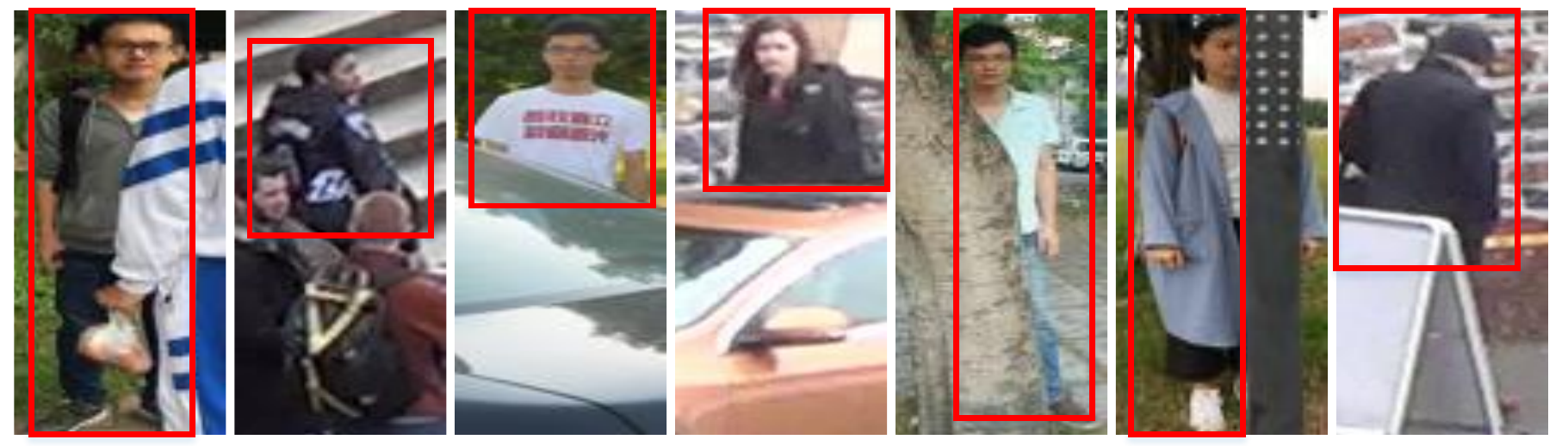}
          \caption{Illustration of occluded person re-id. Red bounding box indicates the target person which is occluded by diversified obstacles with different colors, sizes and positions. }
          \label{fig:occlusion_vis}
          \vspace{-4mm}
        \end{figure}

   Person re-identification (ReID) aims to retrieve a probe pedestrian from non-overlapping camera views. 
   It is an important research topic in computer vision field with various applications, such as autonomous driving, video surveillance and activity analysis~\cite{wojke2017simple, ristani2018features, Li_2019_CVPR}. 
   Most existing ReID approaches design the matching model with the assumption that the entire body of the pedestrian is available.
   However, this assumption is hard to be satisfied due to the inevitable occlusions in real-world scenarios. 
   For example, as shown in Figure.~\ref{fig:occlusion_vis}, a person may be occluded by other pedestrians, static obstacles like trees, walls and cars, etc.
   Therefore, it is essential to seek an effective method to solve this occluded person re-identification problem.

   There are two main challenges for the occluded person ReID task.
     First, the global image-based supervision for conventional person ReID may involve not only the information of the target person but also the interference of occlusion.
     The diversified occlusions, such as colors, positions and sizes, enhance the difficulty of getting a robust feature for the target person.
     Second, the occluded body parts sometimes show more discriminative information while the non-occluded body parts share a similar appearance, leading to the problem of mismatching.

\begin{figure*}
             \centering
             \includegraphics[width=\linewidth]{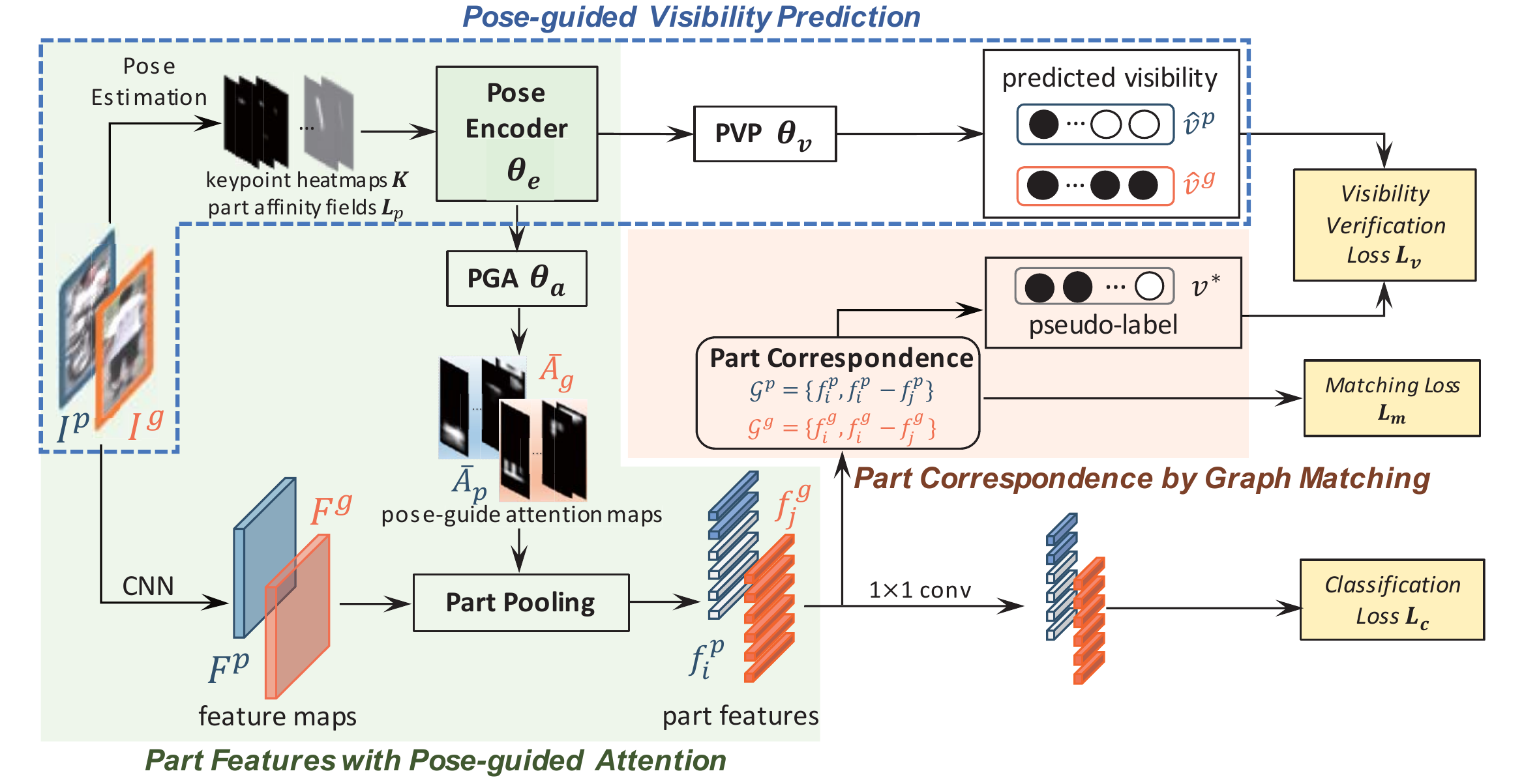}
             \caption{The pipeline of the proposed PVPM approach. It consists of three key components: a pose-guided attention (PGA) model for part feature pooling, a pose-guided visibility predictor (PVP), and a feature correspondence model for providing pseudo-label for the training of PVP. Three loss functions are employed, including $L_v, L_m$ and $L_c$. }
             \label{fig:pipeline}
             \vspace{-4mm}
           \end{figure*}

   An intuitive solution is to detect the non-occluded body parts and then match the correspondents separately. 
   As there is no ground truth annotation for the occluded part, 
   most existing methods directly utilize visibility cues from other tasks with different data source, e.g. body mask~\cite{DBLP:conf/cvpr/CaiWC19} and pose landmark estimation~\cite{miaopose}, but suffering a huge data bias without the flexibility on target domain.
   In this work, we proposed a Pose-guided Visible Part Matching (PVPM) network by directly mining the visible score in a self-learning manner.
   The concept of the proposed approach is demonstrated in Figure~\ref{fig:pipeline}. 
   As shown, PVPM includes two main components: a pose-guided part attention (PGA) network and a pose-guided visibility predictor (PVP) in an end-to-end framework.
   The training of the part visibility predictor is supervised by a pseudo-label obtained by solving a feature correspondence problem via graph matching. 
   In the end, the final score can be computed via the summation of body-part distance aggregation weighted by the visibility score.

      In conclusion, the main contribution of the proposed method is as following:
          \begin{itemize}
       
          \item We propose a Pose-guided Visible Part Matching (PVPM) method that jointly learn the discriminative features with pose-guided attention and predict the part visibility in an end-to-end framework.

          \item We train the visibility prediction model under a self-supervised manner, of which its pseudo-label generating process is regrad as a feature correspondence problem and is solved via graph matching.

         \item The proposed approach achieves superior performance on multiple occlusion datasets including Partial-REID~\cite{zheng2015partial}, Occluded-REID~\cite{zhuo2018occluded} and P-DukeMTMC-reID~\cite{zhuo2018occluded}. 
          \end{itemize}      
   
       \section{Related Work}
       \noindent\textbf{Occluded Person ReID.} 
       Most existing ReID works~\cite{kalayeh2018human,zhang2017alignedreid,DBLP:conf/cvpr/LiuNYZCH18,DBLP:conf/aaai/ZhouFZSLWL18,chang2018multi,li2018harmonious,zheng2015scalable} focus on training the model without taking the occlusions into considerations.
       However, the occlusion can not be ignored especially in the crowd scenes like airports or hospitals.
       To address this problem, Zhou \etal~\cite{zhuo2018occluded} propose multi-task losses that force the network to distinguish between simulated occluded samples and non-occluded samples, so as to learn a robust feature representation against occlusion.
       Besides, a co-saliency network\cite{zhuo2019novel} is proposed to train model paying attention to the person body parts. 
       More recently, Miao \etal~\cite{miaopose} utilize pose landmarks to disentangle the useful information from the occlusion noise. Although the improvement has been made by introducing the pose landmarks, its untrainable pose-guided region extracting and the predefined landmarks visibility still limit the matching performance. 
       Instead of simply using predefined regions and part visibility that learn from other data sources with limited flexibility and data bias, we try to self-mine the part visibility from target data and adapt pose-guide attention accordingly in a unified framework.
           
        \noindent\textbf{Part-based Person ReID.}    
       Part-based person ReID approaches exploit local descriptors from different regions to enhance the discriminative ability and robustness of the algorithm.
       A straightforward way to do this is to slice the person images or feature maps into uniform partitions~\cite{zhang2017alignedreid,sun2018beyond}.
       In~\cite{sun2018beyond}, Sun~\etal partition feature maps into $\mathit{p}$ horizontal stripe and train each part embedding with non-shared classifiers. 
       One can also extract the local features by pose-driven RoI extraction~\cite{zhao2017spindle,su2017pose}, human parsing results~\cite{kalayeh2018human} or learning attention regions based on appearance feature~\cite{li2018harmonious,zhao2017deeply,liu2017hydraplus} or pose feature~\cite{suh2018part}.
       For example, Zhao~\etal~\cite{zhao2017spindle} propose to utilize pose detection results to generate local region by a manual-designed cropping manner, and then fuse the part features gradually. 
       Kalayeh~\etal~\cite{kalayeh2018human} utilize human semantic parsing results to extract body part features.
       Suh~\etal~\cite{suh2018part} propose to generate part maps from prior pose information and then aggregate all parts with a bilinear pooling.
       %
       %
       In~\cite{li2018harmonious, zhao2017deeply,liu2017hydraplus}, they attempt to use appearance-based attention maps to exploit local information. 
       %
       %
       Although the local features are considered in model design, there are no cues for partial occlusion, leading mismatch in the complex environment.
       %

        \noindent\textbf{Self-Supervised Learning.} 
       For the specific task of part visibility prediction, it appears that the precise label of each body part is unavailable to be obtained.
       This motivated us to solve this problem under the self-supervised learning manner.
       Self-supervised learning is proposed to learn feature from unlabelled data by introducing a so-called pretext task, for which a target objective can be computed with self-generated pseudo-label, such as spatial and temporal structure~\cite{jing2018selfsupervised,noroozi2016unsupervised,Wei_2018_CVPR}, or context similarity~\cite{Noroozi_2018_CVPR,Caron_2018_ECCV}.
       Noroozi \etal.~\cite{noroozi2016unsupervised} define the pretext task as recognizing the order of the shuffled patches from an image. Caron \etal.~\cite{Caron_2018_ECCV} treat the cluster assignments as pseudo-labels to learn the parameters of ConvNet.  
       Unlike the pretext task designed above, in this work, we generate the pseudo-label for visibility predictor by self-mine which utilizes the characteristic of part correspondence in positive pairs via graph matching.
          
       \section{Pose-Guide Visible Part Matching}
        
       In this work, we present a pose-guided visible part matching framework which aggregates local features with the visible scores to solve the mismatching problem for occluded person ReID task. 
       To better understand the proposed method, we demonstrate the pipeline and the training process in Figure~\ref{fig:pipeline} and Algorithm Box~\ref{alg:A}, with the related notations illustrated in Table~\ref{tab:notation}. 
       This framework includes a pose encoder (PE), a pose-guide attention mask generator (PGA), a pose-guided visibility score predictor (PVP) and a feature correspondence model for generating the pseudo-label for training PVP. 
       In Sec.~\ref{PGA} and~\ref{PVP}, we introduce the methodology details of the PGA and PVP modules.
       In Sec.~\ref{PCG}, we claim the strategy about how to obtain the pseudo-label of part correspondence to supervise the training of PVP.
       Last, in Sec.~\ref{loss-function}, we demonstrate the formulation of the loss functions we employed in this method.

       \subsection{Part Features with Pose-Guide Attention}\label{PGA}
       Obviously, discriminative part features play an important role for the circumstance when the target person facing with occlusions.
       This motivates us to get the body part features by fusing the appearance features with pose-guided attention maps.
       With a given pedestrian image $I$, we first extract the appearance feature maps $F \in\mathbb{R}^{C\times H\times W}$ via a CNN backbone network,
       where C, H, W denote the number of pixel in the channel, height, and width dimensions of feature maps, respectively. 

       The pose-guided attention mechanism consists of three components: pose estimation, pose encoder and part attention generator.
       We employ the Openpose~\cite{cao2017realtime} method for pose estimation to extract the key point heatmaps ${K}$ and the part affinity fields ${L_{p}}$ of each input image. 
       The pose encoder then takes $P = K\oplus L_{p}$ as input and embeds the pose information into a high-level pose feature $F_{pose}=PE(P;\theta_{e})$.
       For the part attention generator which focuses on a specific body part, a $1 \times 1$ Convolutional layer and a following Sigmoid function is adopted on pose features $F_{pose}$ to estimate a stack of 2-dimensional maps $A$, each element $a_{i}^{h,w}$ in $A$ indicates the degree that the location $(h,w)$ from feature maps $F$ lies in the $i$-th part:
           \begin{equation}\label{eq:part_maps}
             A=PGA(F_{pose};\theta_a)\in \mathcal{R}^{{N_p}\times H\times W}
           \end{equation}
       where $N_p$ is the number of pre-defined parts, $\theta_a$ is the parameters of the convolutional layer. 
       Furthermore, we hope the network could focus on the nonoverlapping region so that each part could extract complementary features which are more discriminative and robust when fusing them all. Thus, we only maintain the maximum activation along the first channel for each part map, which is formulated by,
           \begin{equation}
             \bar{A}_{i}=A_i\odot [\mathop{\arg\max_i}A_{i}]|^C_{onehot}
           \end{equation}
       $\odot$ is the Hadamard Product, $[\mathop{\arg\max_i}A_{i}]|^C_{onehot}$ means to get the index of maximum value along the channel dimension and turn it into a one-hot vector at each spatial location.
       
       In the end, the $i$-th part feature $f_i$ can thus be obtained via a part weighted pooling, which is formulated by,
           \begin{equation}\label{eq:part_feat}
             f_i=\frac{1}{\|\bar{A_i}\|}\sum^{H}_{h=1}\sum^{W}_{w=1}\bar{a}^{h,w}_{i}\odot F^{h,w}
           \end{equation}
           \begin{equation}
             \|\bar{A}_i\| = \sum^{H}_{h=1}\sum^{W}_{w=1}\bar{a}_i^{h,w}
           \end{equation}
       where $F^{h,w}$ is the column vector of $F$ at position $(h,w)$, $\bar{a}_i^{h,w}$ denotes the element lies in the location (h,w) of $\bar{A}_i$.
       %

       \subsection{Pose-Guide Visibility Prediction}\label{PVP}
       After representing the pedestrian using part-based features, an intuitive way to calculate the distance is to compute the global part-to-part distances.
       However, for occluded ReID, some patches appear in one view may not be exposed in other views.
       Therefore, a reasonable way is to only establish the correspondence between simultaneously visible parts and compute the distance accordingly.
       %
       %
       We propose to utilize a pose-guide visibility score predictor (PVP) to estimate the visibility for each part.
       %
       %
       
       We implement the PVP method via a four-layer tiny network which consists of a global average pooling layer, a Convolutional layer of $1 \times 1$ filter, a BatchNorm layer and a Sigmoid activation layer.
       With an input pose feature $F_{pose}$, we can predict the visibility score through,
           \begin{equation}\label{eq:vis_score}
             \hat{v}=PVP(F_{pose}; \theta_v)\in \mathcal{R}^{N_p}
           \end{equation}

       When it comes to the testing stage, given a probe image $I_p$ and a gallery image $I_g$, the distance considering the visibility between them can be calculated as:
          \begin{equation}
             d=\frac{\sum^{N_p}_{i=1}\hat{v}_{i}^{p}\hat{v}_{i}^{g}d_i}{\sum^{N_p}_{i=1}\hat{v}_{i}^{p}\hat{v}_{i}^{g}}
          \end{equation}
       where $d_i$ is the cosine distance of the $i$-th part features, $\hat{v}^{i}_{p}$ and $\hat{v}^{i}_{g}$ denote the visibility score of the $i$-th part of the probe image and gallery image, respectively.    
        
         \begin{table}
             \caption{Notation definition.} 
             \centering
             \small
             \label{tab:notation}
             \scalebox{1}
             {\begin{tabu}{cc}
            \hline
           $I^p, I^g$ & probe/gallery images\\
           \hline
           $F^p, F^g$ & probe/gallery feature extracted by CNN\\
           \hline
           $F_{pose}$ & pose features after pose encoder\\
           \hline
           $\bar{A}^p, \bar{A}^g$ & pose-guide attention maps of probe/gallery \\
           \hline
           $f_i^p, f_j^g$ & $i/j$-th part features of probe/gallery\\
           \hline
           $\mathcal{G}^p, \mathcal{G}^g, M$ & graph of probe/gallery and its affinity matrix\\
           \hline
           $\hat{v}^p,\hat{v}^g $ & predicted visibility of probe/gallery via PVP\\
           \hline
           $v^*$ & visibility optimized by correspondence learning \\
           \hline
           $\theta_e, \theta_v, \theta_a$ & parameter of pose encoder, PVP, PGA \\
           \hline
            $\lambda$ & regularization coefficient\\
            \hline
            $T$ & maximal iteration for training \\
            \hline
            $N_p$ & parts number for each image\\
            \hline
             \end{tabu}
             }
          \end{table}

       \subsection{Pseudo-Label Estimation by Graph Matching}\label{PCG}
       The ground-truth visibility label of each part is usually unavailable. 
       This motivates us to seek a method that can automatically reveal the visible part without further requirement of manually occlusion annotation in a self-supervised way.
       For a given positive image pair $I^p, I^g$, we observe that (1) the relevance of a part-pair appears to be high only when both parts in $I^p, I^g$ are visible. 
       (2) the relevance between the edges of two visible parts within the image will also be highly correlated.

\begin{figure}
             \centering
             \includegraphics[width=\linewidth]{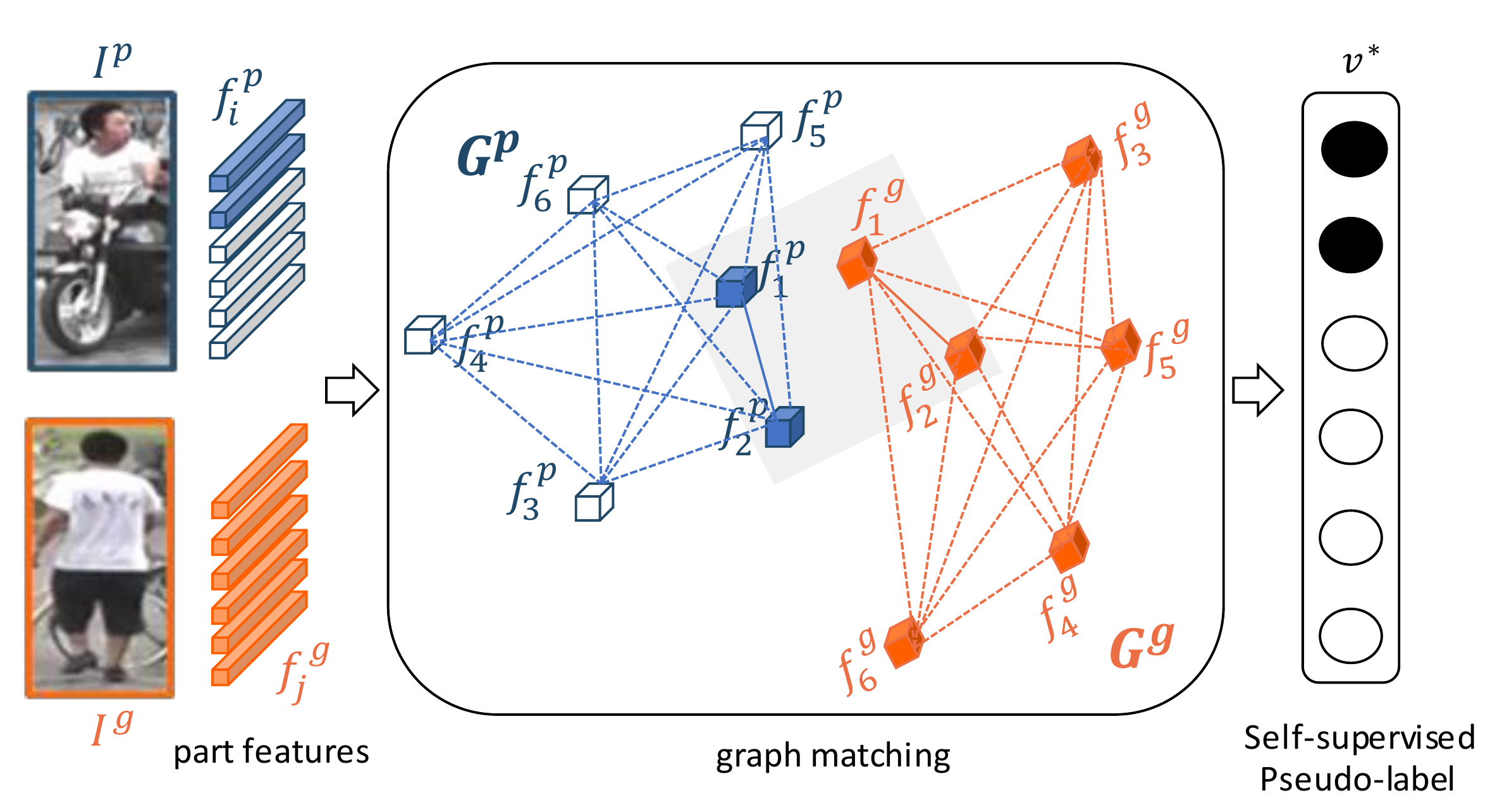}
             \caption{Pseudo-label estimation via graph matching.}
             \label{fig:cor_learn}
           \end{figure}

       Based on these observations, instead of training $\hat{v}$ directly, we train the product of part visibility scores of positive pairs to approximate their correspondence. 
       Thus, we consider the pseudo-label generation process as a part feature correspondence problem which can be solved by graph matching.
       %
       %
       %
       For better understanding, we illustrate an example of how to obtain the pseudo-label between two input images in Figure~\ref{fig:cor_learn}.
       %
       %
       
       Specifically, for a given positive pair, we represent them via two graphs $\mathcal{G}^p = (\mathcal{V}^p,\mathcal{E}^p)$ and $\mathcal{G}^g = (\mathcal{V}^g,\mathcal{E}^g)$, where each element $\mathcal{V}_{i}$ and $\mathcal{E}_{i,j}$ denote the parts(nodes) features {$f_i$} and edges features $\{f_i-f_j\}$ respectively.
       In our task, only one-to-one matching between corresponding nodes of two graphs are adopted. A binary indicator vector $v \in \{0,1\}^{N_P}$ is employed to represent the correspondence of the two input parts from $\mathcal{G}^p$ and $\mathcal{G}^g$, where $v_i$ set as 1 if the $i$-th part pair is selected for matching, otherwise 0. 
       The affinity matrix $M$ is conducted with the relational similarity values between edges and nodes where the inner product is used to calculate similarity. 
       %
       %
       Specifically, we encode the compatibility of corresponding two nodes in the diagonal $M_{ii}$ as:
          \begin{equation}
             M_{i,i}=\langle f^{p}_{i},f^{g}_{i}\rangle
          \end{equation}
          and encode the compatibility of corresponding two edges features in the non-diagonal component $M_{ij}$ as:
          \begin{equation}
             M_{i,j}=\langle \frac{\mathcal{E}^p_{i,j}}{\left\|\mathcal{E}^p_{i,j}\right\|_2},\frac{\mathcal{E}^g_{i,j}}{\left\|\mathcal{E}^g_{i,j}\right\|_2}\rangle-\hat{M}_{i,j}
          \end{equation}
          where $\hat{M}_{i,j}$ is the moving average of $M_{i,j}$.
       
          Same as another graph matching method~\cite{suh2015subgraph}, we model graph matching as an Integer Quadratic Programming (IPQ) problem and incorporate a regularization term on the number of activated nodes:
          \begin{equation}\label{eq:solve_v}
               \mathop{\arg\max_{v}} \quad v^TMv-\bar{\lambda}^T v
               \quad \mbox{s.t.} \  v\in\{0,1\}^{N_p}.
          \end{equation}
          \begin{equation}
             \bar{\lambda}=\lambda\hat{M}_{diag}
          \end{equation}
         where $\lambda$ is a balanced parameter and $\hat{M}_{diag}$ is the moving average of diagonal components of $M$. 
         We set $\bar{\lambda}$ to be proportional to the moving average of parts similarity to make it more adaptive to data as well as narrow down the scope of hyper-parameter selection.
       By optimizing Eq.(\ref{eq:solve_v}), we can obtain the optimal solution $v^*$ which indicates which part pair is appropriate to be matched. 
       Then it can be taken as the supervision for optimizing PVP. 
       
       \subsection{Loss Function}\label{loss-function}
       Three loss functions are employed to optimize the proposed method, including the visibility verification loss $L_v$ for self-supervised visibility learning, the part-matching loss $L_m$ for enhancing the relevance between corresponding parts, and the identity classification loss $L_c$ for maintaining the discriminative power of each part feature.
       Therefore, the overall loss $L$ can be formulated as,
        \begin{equation}
               L=L_v+L_c+L_m
        \end{equation}
        
       {\noindent\textbf{Visibility Verification Loss $L_v$. }} We impose a Binary Cross Entropy loss for PVP module in training phrase with the self-supervision signal $v^*$, which is obtained via strategy mentioned in Sec.~\ref{PCG}. 
       Specifically, the product of part visibility scores of the input probe and gallery $I^p, I^g$ are trained to approximate matching vector, which is formulated by:
          \begin{equation}
             L_v = -\sum^{N_p}_{i=1}v^*_{i}\log(\hat{v}_{i}^{p}\hat{v}_{i}^{g})
          \end{equation}
          where $v_{i}^{p}$ and $v_{i}^{g}$ correspond to the $i$-th part visibility score of probe and gallery images respectively.
       
       {\noindent\textbf{Part Matching Loss $L_m$. }}  
       After obtaining the optimal visibility score $v^*$, continue to optimize the matching quality function according to $M$ enables to enhance the intra-part consistence. 
       A part-based matching loss which is similar to the form of Eq.\ref{eq:solve_v} is employed here. 
       By fixing $v$ with value $v^*$, the matching loss is formulated as:
          \begin{equation}
             L_m = -v^{*T}Mv^*+\lambda'^Tv^*
          \end{equation}
       Among this loss function, the first term could enhance the intra-part consistence and the second term enforces the network to extract complementary features from different parts, where $\lambda'\in \mathcal{R}^{P}$ is defined as,
          \begin{equation}
             \lambda'_i= \frac{\sum^{N_p}_{j=1,j\ne i}S_{ij}^p+S_{ij}^g}{2({N_p}-1)}
          \end{equation}
          and $S^p$ and $S^g$  corresponding to the inter-part feature similarity matrix of probe and gallery, respectively.   
       
       {\noindent\textbf{Classification Loss $L_c$. }} 
       To introduce discriminative power into the proposed network, we adopt a classification loss as the objective function.
       Following the construction of RPP~\cite{sun2018beyond}, we fix the pre-trained PCB classifiers to maintain the knowledge learned under the uniform partition. 
       Then the classification loss can be formulated as,
          \begin{equation}
             L_c = \sum^{N_p}_{i=1}CE(\hat{y}_i,y_i)
          \end{equation}
          where $CE$ is the Cross-Entropy loss, $\hat{y}_i$ is the prediction of the $i$-th part classifier, and $y$ is the ground-truth ID.
       
          \begin{algorithm}[t]
             \caption{Pose-Guided Visible Part Matching. }
             \renewcommand{\algorithmicrequire}{\textbf{Input:}}
             \renewcommand{\algorithmicensure}{\textbf{Output:}}
             \label{alg:A}
             \begin{algorithmic}[1]
                \REQUIRE Training image data: $\textbf{I}$, $T$, $\lambda$.
                \ENSURE The parameters $\theta_e$, $\theta_v$, $\theta_a$ of PE , PVP and PGA 
                \STATE Initialize $\theta_e$,$\theta_v$, and $\theta_a$
                \FOR{$t=1,2,\dots, T$}
                \STATE Randomly select a batch of images from \textbf{I};
                \STATE Generate part feature $\{f_i\}$ with Eq.~\ref{eq:part_feat}
                \STATE Predict visibility score $\{\hat{v}_i\}$ with Eq.~\ref{eq:vis_score}
                \STATE Obtain pseudo label $\{v^*_i\}$ by solving Eq.\ref{eq:solve_v}
                \STATE Update $\theta_v \gets \frac{\partial}{\partial\theta_v}L_v$ 
                \STATE Update $\theta_a \gets \frac{\partial}{\partial\theta_a}(L_c+L_m)$ 
                \STATE Update $\theta_e \gets \frac{\partial}{\partial\theta_e}(L_v+L_c+L_m)$
                \ENDFOR
                \RETURN $\theta_v$,$\theta_a$, and $\theta_e$
             \end{algorithmic}
          \end{algorithm}
         
           \begin{table*}
           \renewcommand\arraystretch{1}
             \caption{
            Performance comparisons with the holistic and occluded methods on the three reported datasets. The 1\textsuperscript{st}/2\textsuperscript{nd} best results are in \textcolor{red}{red} and \textcolor{blue}{blue}.}
             \label{tab:scores}
             \centering
             \scalebox{.93}
             {
             \begin{tabular}{ccccccccccccc}
             \hline
              \multirow{2}{*}{Method}  &\multicolumn{4}{c}{Occluded-REID}&\multicolumn{4}{c}{Partial-REID}&\multicolumn{4}{c}{P-DukeMTMC-reID}\\
              \cmidrule(r){2-5} \cmidrule(r){6-9} \cmidrule(r){10-13}
                   & rank-1 & rank-5 & rank-10 & mAP  & rank-1 & rank-5 & rank-10 & mAP & rank-1 & rank-5 & rank-10 & mAP  \\
              \hline
              IDE~\cite{zheng2015scalable} &52.6 &68.7 &76.6 &46.4 &51.7 &69.0 &80.3 &52.4 &36.0 &49.3 &55.2 &19.7\\
              OsNet~\cite{zhou2019omni} &39.7 &57.9 &66.5 &36.0 &48.7 &68.0 &78.3 &49.3 &33.7 &46.5 &54.0 &20.1\\
              MLFN~\cite{chang2018multi} &42.3 &60.6 &68.5 &38.4 &42.7 &62.7 &72.3 &45.7 &31.3 &43.6 &49.6 &18.1\\
              HACNN~\cite{li2018harmonious} &29.1 &44.7 &54.7 &26.1 &37.0 &64.0 &75.3 &40.4 &30.4 &42.1 &49.0 &17.0\\
              Part Bilinear~\cite{suh2018part} &54.9 &70.8 &77.7 &50.3 &57.7 &77.3 &85.7 &59.3 &39.2 &50.6 &56.4 &25.4\\
              PCB~\cite{sun2018beyond} &59.3 &75.2 & 83.2 &53.2 & 66.3 &84.0 &91.0 & 63.8 &43.6 &57.1 &63.3 &24.7\\
              PCB+RPP~\cite{sun2018beyond}  &55.8 &74.4 &81.2 &51.3 &63.7 &82.3 &90.0 &61.2 &40.4 &54.6 &61.1 &23.4\\
             \hline
              Teacher-S~\cite{zhuo2019novel} &55.0 &64.5 &77.3 &\textcolor{red}{59.8} &69.2 & 76.6 &85.8 &\textcolor{red}{73.1} &18.8 &24.2 &32.2 &22.4\\
              PGFA~\cite{miaopose} &57.1   &77.9 &84.0 &56.2 &68.0 &82.0 &86.7 &56.2 &44.2 &56.7 &63.0 &23.1\\
            \hline
                PVPM  &\textcolor{blue}{66.8} &\textcolor{blue}{82.0} &\textcolor{blue}{88.4} &59.5 &\textcolor{blue}{75.3} &\textcolor{blue}{88.7} &\textcolor{blue}{92.3} &71.4 &\textcolor{blue}{50.1} &\textcolor{blue}{63.0} &\textcolor{blue}{68.6} &\textcolor{red}{29.4}\\
                PVPM+Aug &\textcolor{red}{70.4} &\textcolor{red}{84.1} &\textcolor{red}{89.8} &\textcolor{blue}{61.2} &\textcolor{red}{78.3} &\textcolor{red}{89.7} &\textcolor{red}{93.7} &\textcolor{blue}{72.3} &\textcolor{red}{51.5} &\textcolor{red}{64.4} &\textcolor{red}{69.6} &\textcolor{blue}{29.2}\\
            \hline
              \end{tabular}
             }
             \vspace{-3mm}
             \end{table*}

       \section{Experiments}
       \subsection{Datasets and Settings}
       \noindent{\textbf{Datasets. }}
       For experimental evaluation, we conduct experiments on two small-scale and one large-scale ReID benchmarks, including the Occluded-REID~\cite{zhuo2018occluded}, the Partial-REID~\cite{zheng2015partial}, and the large P-DukeMTMC-reID~\cite{zhuo2018occluded} dataset. 
       Each reported occluded dataset is partitioned into two parts: the occluded person images and full-body person images.
       For model pre-training, we train the networks on the {Market-1501}~\cite{zheng2015scalable} dataset.  
       
       1) {Occluded-REID}~\cite{zhuo2018occluded} images are captured by mobile camera equipments in campus, including 2,000 annotated images belonging to 200 identities. 
       Among the dataset, each person consists of 5 full-body person images and 5 occluded person images with various occlusions.
       
       2) {Partial-REID}~\cite{zheng2015partial} includes 900 images of 60 pedestrians. 
       Each person has 5 full-body person images, 5 occluded person images and 5 manually cropped partial person images from the occluded ones. 
       In this work, we only use the full-body and occluded person images for evaluation.

       3) {P-DukeMTMC-reID}~\cite{zhuo2018occluded} is a modified version based on DukeMTMC-reID dataset~\cite{zheng2017unlabeled}. 
       There are 12,927 images (665 identifies) in training set, 2,163 images  (634 identities) for querying and 9,053 images in the gallery set.  
       
       4) {Market-1501}~\cite{zheng2015scalable} contains 32,668 labelled images of 1,501 identities observed from 6 cameras.
       The dataset is split into training set with 12,936 images of 751 identities and used for model pre-training only.
       
       \noindent{\textbf{Evaluation Protocols. }}
       We report the Cumulated Matching Characteristics (CMC)~\cite{gray2007evaluating} and mean Average Precision (mAP)~\cite{zheng2015scalable} value for the proposed approach.
       The evaluation package is provided by \cite{torchreid}, and all the experimental results are performed in a single query setting.

       \noindent{\textbf{Implementation Details}}
       We take all occluded person images as probe set and full-body person images as gallery set on all three reported datasets. 
       Specifically, for the {Occluded-REID}~\cite{zhuo2018occluded} and {Partial-REID}~\cite{zheng2015partial} datasets, due to the absence of the same prescribed split of training and test set, all the images are adopted for testing. 
       With all training images resized as $384 \times 128$, we employed ResNet-50~\cite{he2016deep} which is pre-trained with the same setting as PCB~\cite{sun2018beyond} to extract appearance features.
       This feature is then followed by a pose-guided attention pooling operation which generates $N_p$ part features, where $N_p$ is set as 6 by default.
       For pose estimation, we adopt the OpenPose~\cite{cao2017realtime} method pre-trained on the COCO dataset~\cite{lin2014microsoft}, which generates 18 keypoint heatmaps $K$ and 38 part affinity fields $L_p$.
       The proposed PVP and PGA method is trained at a learning rate of 0.002 via the SGD optimizer.
       The training batch size, the training epoch, and the coefficient $\lambda$ are set as 32, 30, and 0.9, respectively.
       This code is implemented under NVIDIA 1080Ti GPU environment and Pytorch platform.
        
       \subsection{Performance under Transfer Setting}\label{sec:transfer}
       Performance comparison under transfer setting is conducted by directly utilizing the model trained on Market1501~\cite{zheng2015scalable} without any further optimization.
       
       \noindent\textbf{Comparison with Holistic Methods}
       The performance comparison with the holistic methods are illustrated in the first group of Table ~\ref{tab:scores}.
       Among these methods, HACNN~\cite{li2018harmonious} introduces the appearance-based attention mechanism into model training. 
       Compared to the Part Bilinear~\cite{suh2018part} method which utilizes the pose information to improve the re-identification performance, the PCB(+RPP)~\cite{sun2018beyond} method propose to use a  refined part pooling strategy.
       The `+Aug' corresponds to the result when training the PVPM model with images augmented with random occlusions to solve the data unbalance problem in the occluded training set.
       From the table, we can observe that the proposed method outperforms those holistic approaches by a large margin, with rank-1 accuracy surpasses the second-best holistic method by around 10\% for all three reported benchmarks.
       This result may validate that,
       1) it is essential to propose a specifically designed framework for the occluded ReID task;
       2) matching with the visible parts shows better performance rather than using all parts.
       %
           \begin{table}
           \renewcommand\arraystretch{1}
             \caption{Comparison with partial ReID methods on the Partial-REID dataset. The `manually crop' indicates the method use the original occluded images or the manually occlusion removed images for matching.}
             \centering
             \label{tab:partial}
             \scalebox{.93}
             {\begin{tabu}{cccc}
            \hline
             Method & rank-1 & rank-3   & manually crop\\
             \hline
             MTRC~\cite{liao2012partial} &23.7 &27.3    & $\surd$  \\
             AWC+SWM~\cite{zheng2015scalable} &37.3 &46.0  &$\surd$   \\
             SFR~\cite{he2018recognizing} &56.9 &78.5 & $\surd$\\
             VPM~\cite{sun2019perceive} &67.7 &81.9    & $\surd$\\
             \hline
             PVPM  &75.3 &86.0  & $\times$\\
            PVPM+Aug &\textcolor{red}{78.3} &\textcolor{red}{87.7}
     &  $\times$ \\
             \hline
             \end{tabu}
             }
          \end{table}
               \begin{table}
             \renewcommand\arraystretch{1}
             \caption{Performance on the P-DukeMTMC-reID dataset under supervised setting.} 
             \centering
             \label{tab:supervised}
             \scalebox{.93}
             {\begin{tabu}{ccccc}
            \hline
             Method & Rank-1 & Rank-5 &Rank-10 &mAP \\
             Teacher-S~\cite{zhuo2019novel} &51.4 &50.9 &- &-\\
             IDE~\cite{zheng2015scalable} &82.9 &89.4 &91.5 &65.9\\
             Baseline(PCB)~\cite{sun2018beyond} &79.4 &87.1 &90.0 &63.9      \\
             PVPM &\textcolor{red}{85.1} &\textcolor{red}{91.3} &\textcolor{red}{93.3} &\textcolor{red}{69.9}   \\
            \hline
             \end{tabu}
             }
             \vspace{-3mm}
          \end{table}   
           
       \noindent\textbf{Comparison with Occluded Methods}
       We show the performance comparison with two specifically designed occluded ReID approaches in the second group of Table ~\ref{tab:scores}.
       The Teacher-S~\cite{zhuo2019novel} proposes to train networks to learn a global feature with two auxiliary tasks, which would make networks paying more attention to person body parts.
       The PGFA~\cite{miaopose} proposes a hard part matching method via a fixed region selection strategy and hand-crafted part visibility judgement method.
       Compared to these two approaches, our PVPM model achieves 70.4\%, 78.3\% and 51.5\% at rank-1 on the Occluded ReID~\cite{zhuo2018occluded}, {Partial-REID}~\cite{zheng2015partial}  and {P-DukeMTMC-reID}~\cite{zhuo2018occluded} dataset, outperforming them by a large margin.
       This large performance improvement may be drawn from three aspects:
       1) part matching works better for occluded ReID task rather than global feature learning;
       2) a trainable part visibility prediction model could benefit more than the hand-crafted strategy;
       3) training a high-level pose features can provide better guidance for person retrieval compared to simply fuse features with pose keypoint heatmaps;
          
        \begin{table*}
          \renewcommand\arraystretch{1}
                \caption{Performance comparisons with different component settings. }
                \label{tab:ablation study}
            
                \centering
                \scalebox{.93}
                {
                \begin{tabular}{ccccccccccccc}
                 \hline
                 \multirow{2}{*}{Method}  &\multicolumn{4}{c}{Occluded-REID}&\multicolumn{4}{c}{Partial-REID}&\multicolumn{4}{c}{P-DukeMTMC-reID}\\
                 \cmidrule(r){2-5} \cmidrule(r){6-9} \cmidrule(r){10-13}
                      & rank-1 & rank-5 & rank-10 & mAP  & rank-1 & rank-5 & rank-10 & mAP & rank-1 & rank-5 & rank-10 & mAP  \\
                      
                 \hline
                Baseline(PCB)~\cite{sun2018beyond} &59.3 &75.2 & 83.2 &53.2 & 66.3 &84.0 &91.0 & 63.8 &43.6 &57.1 &63.3 &24.7\\
               PVPM &66.8 &82.0 &\textcolor{blue}{88.4} &59.5 &\textcolor{blue}{75.3} &88.7 &\textcolor{blue}{92.3} &\textcolor{blue}{71.4} &\textcolor{blue}{50.1} &63.0 &68.6 &\textcolor{red}{29.4}\\
               PVPM-$L_m$ &65.5 &80.7 &86.3 &58.2 &73.0 &86.7 &92.0 &69.6 &48.1 &61.7 &67.9 &27.7\\
               PVPM-thre &65.1 &80.3 &87.3 &58.1 &71.7 &87.0 &91.0 &68.1 &48.1 &61.5 &68.1 &29.0\\
                PGA only &61.1 &77.2 &84.5 &55.0 &68.7 &85.0 &91.7 &65.6 &43.9 &58.1 &64.5 &26.6 \\
                PVP only &65.2 &80.4 &86.6 &57.3 &74.0 &\textcolor{blue}{89.3} &\textcolor{blue}{92.3} &70.4 &46.8 &62.0 &67.4 &26.0 \\   
                PVP only+Aug &\textcolor{blue}{69.0} &\textcolor{blue}{83.5} &\textcolor{blue}{88.4} &\textcolor{blue}{60.9} &74.7 &89.0 &\textcolor{blue}{92.3} &71.2 &49.7 &\textcolor{blue}{63.3} &\textcolor{blue}{69.3} &27.5\\
                PVPM+Aug &\textcolor{red}{70.4} &\textcolor{red}{84.1} &\textcolor{red}{89.8} &\textcolor{red}{61.2} &\textcolor{red}{78.3} &\textcolor{red}{89.7} &\textcolor{red}{93.7} &\textcolor{red}{72.3} &\textcolor{red}{51.5} &\textcolor{red}{64.4} &\textcolor{red}{69.6} &\textcolor{blue}{29.2}\\

                \hline
                \end{tabular}
                }
              \vspace{-3mm}
                \end{table*} 
          
       \noindent\textbf{Comparison with Partial Methods}
       Compared to occluded ReID task, partial ReID aims to solve the matching problem with the images manually cropped via a bounding box from the original images.
       This may result in image distortion and misalignment, and the occlusions still can not be totally removed, therefore, increasing the matching difficulties.
       In this section, four partial ReID methods are compared with the proposed PVPM in Table~\ref{tab:partial} on the Partial ReID dataset~\cite{zheng2015partial}, listing the rank-1, rank-3 matching rates.
       We also demonstrate whether the model needs to match persons with the manually occlusion removed images or the original pictures.
       As can be seen, compared to those partial ReID methods, our PVPM+Aug model arrives 78.3\% at rank-1, outperforming the second-best VPM~\cite{sun2019perceive} approach by 10.6\%.
       %
       %
       Note that, our PVPM approach does not require to pre-process the images while testing, which shows better practicability in the real-world scenes.

       \subsection{Performance under Supervised Setting}
       For the large-scale dataset P-DukeMTMC-reID~\cite{zhuo2018occluded}, we further run experiments to evaluate the performance when optimizing the model with the target training set. 
       The results of two methods, IDE~\cite{zheng2015scalable} as well as our part-based baseline method PCB~\cite{sun2018beyond} are demonstrated in Table~\ref{tab:supervised}. 
       As can be observed, our PVPM method achieves 85.1\% at rank-1, which surpasses the baseline method by 5.7\%.
       This further illustrates our model superiority under the supervised setting for occluded person ReID. 
       %
       %
       %
           
       \subsection{Algorithm Analysis}
       In this subsection, we conduct experiments to thoroughly verify the effectiveness of the components of the Pose-Guided Attention (PGA) mechanism, the Pose-guided Visibility Prediction (PVP) model, the part matching loss $L_m$, the graph matching model and the augmented training samples with randomly generated occlusions. 
       The experimental results on the reported three benchmarks are shown in Table~\ref{tab:ablation study}.
       The `Baseline' is the result of directly employing the PCB~\cite{sun2018beyond} model.
       The `PGA only' means that we only use the pose-guided part features without further employment of part visibility computation.
       The `PVP only' corresponds to the result that assigning each uniform part features with a visibility score without the soft pose-guided attention mask. 
       The `-$L_m$' is the result when removing the part matching loss from the whole loss function.
       The `-thre' is the result when inferring pair visibility by thresholding their similarity.
       The `+Aug' indicates that we augment the training samples by randomly replacing a region in the image with a background patch, which is motivated by~\cite{zhuo2018occluded}.
       %
        
       From Table~\ref{tab:ablation study}, we can observe that the employment of PGA block can achieve better performance.
       This suggests that the utilization of pose-guide attention do benefit the occluded re-identification task.
       %
       When comparing the result of `PVP only' and `baseline', it can be easily drawn that computing a weighted distance according to the part visibility score improves the rank-1 performance by 5.9\%, 7.7\%, 3.2\% on the three reported datasets. 
       Note that, our graph model method outperforms the thresholding method, which demonstrates our model advantage as it considers the body part-to-part correlations while inferring their correspondence.
       What is more, when we remove the $L_m$ from the entire loss function, performance drops by around 1-2\% at rank-1 accuracy, validating its effectiveness.
       Besides, the result of the `+Aug' operation demonstrates that the augmented occluded training samples can make a contribution to performance gain.

      \begin{table}
         \renewcommand\arraystretch{1.05}
         \caption{Comparison results of generating part maps and visibility score from difference type of cues: appearance-based or pose-guided. 
         PVPM is the proposed pose-guide method, `RPP' indicates to refine the part maps from uniform spliting as in~\cite{sun2018beyond}. `R+S' means the result when we further employ an appearance-based visibility predictor with the `RPP'. }\label{tab:appearance}
         \centering
         \scalebox{0.9}{
         \begin{tabu}{c|ccccc}
             \hline
             Datasets  &Methods & Rank-1 & Rank-5 & Rank-10 & mAP \\
             \hline
             \multirow{3}{*}{Occluded}   & RPP &55.8 &74.4 &81.2 &51.3       \\
             & R+S &51.8   &69.3   &76.6   &47.3     \\
             & PVPM &\textcolor{red}{66.8}   &\textcolor{red}{82.0}   &\textcolor{red}{88.4}   &\textcolor{red}{59.5}     \\
             \hline
             \multirow{3}{*}{Partial}   & RPP &63.7  &82.3  &90.0  &61.2  \\
             & R+S &59.7   &81.3   &88.0   &59.0     \\
             & PVPM &\textcolor{red}{75.3}   &\textcolor{red}{88.7}   &\textcolor{red}{93.7}   &\textcolor{red}{72.3}     \\
             \hline
             \multirow{3}{*}{P-Duke}   & RPP &40.4  &54.6   &61.1  &23.4       \\
             & R+S &35.6   &47.9   &53.3   &21.1     \\
             & PVPM &\textcolor{red}{50.1}   &\textcolor{red}{63.0}   &\textcolor{red}{68.6}  &\textcolor{red}{29.4}     \\
             \hline
         \end{tabu}}
         \label{tab:pose_cues}
         \vspace{-5mm}
      \end{table}
      \begin{figure}[tbp]
         \centering
         \includegraphics[width=\linewidth]{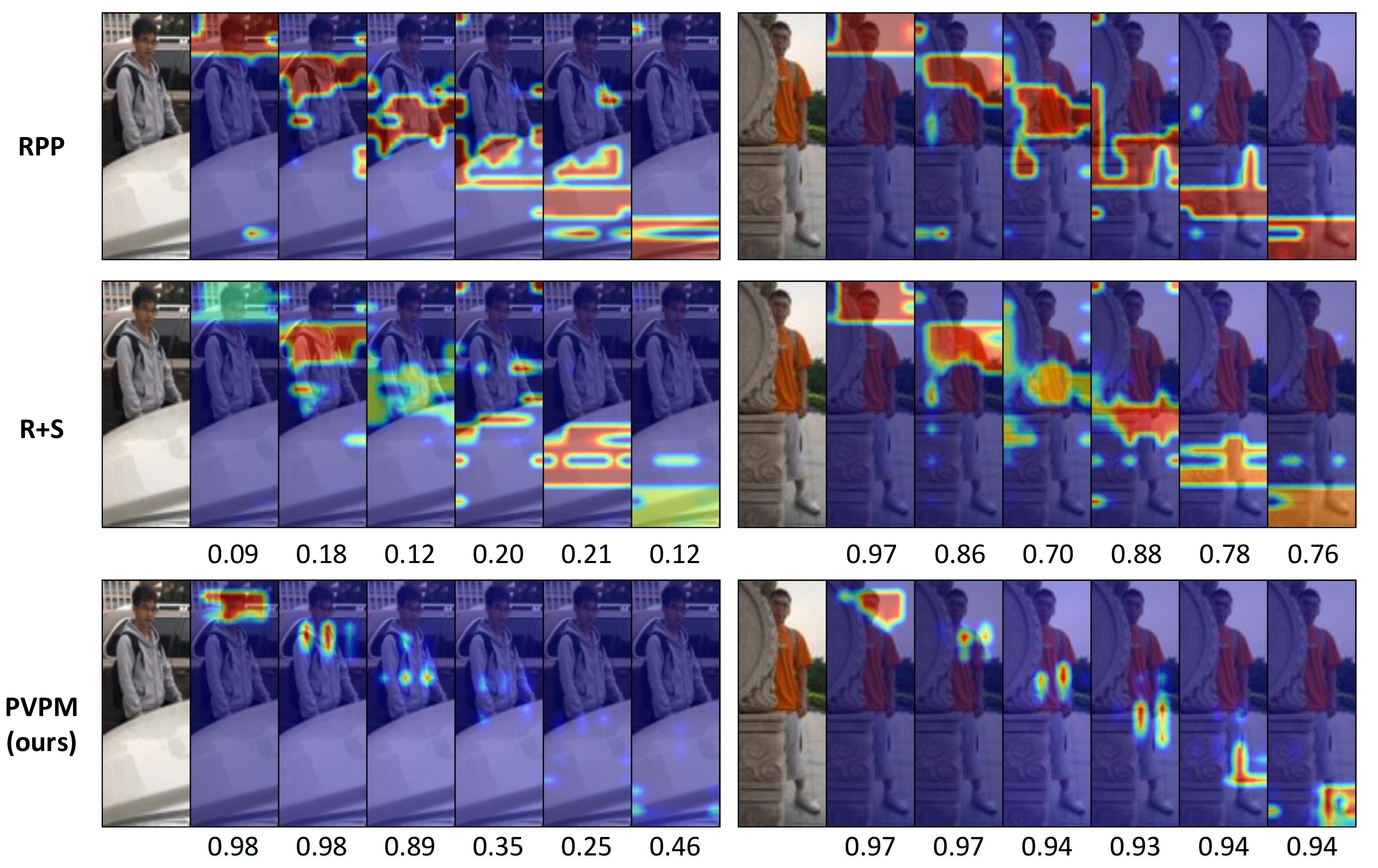}
         \caption{Visualization of part maps and visibility score generated from difference cues. The number under each picture denotes the predicted visibility score of this part. 
         The pictures on each line indicate the part maps generated via our PVPM model, the appearance-based refined part model(RPP~\cite{sun2018beyond}), and the combination of RPP and a visibility score predictor(VSP), respectively}
         \label{fig:advantage_pose}
         \vspace{-3mm}
      \end{figure}
          
       \subsection{Analysis of Pose Cues}\label{sec:advan_pose}
       Compared with the appearance cues, the pose-guided cues sometimes can provide more reliable information for occluded occasions.
       To validate the advantage of utilizing human pose information for part region generation and part visibility score prediction, we compare our PVPM model with an appearance-based part refine method RPP~\cite{sun2018beyond}. 
       The quantitative result is illustrated in Table~\ref{tab:pose_cues}.
       As can be seen, the appearance-based RPP~\cite{sun2018beyond} method does not achieve a performance boost on the occluded dataset comparing with the baseline method PCB~\cite{sun2018beyond} (in Table.\ref{tab:ablation study}).
       %
       Furthermore, we train a model with the same setting as our self-supervised framework but replace the PGA and PVP block with two appearance-based module, RPP~\cite{sun2018beyond} and a part visibility score predictor (VSP), respectively. 
       This strategy is defined as `R+S'. 
       The further employment of the VSP method makes the performance drop further. 
       For better viewing, we demonstrate the visualization result in Figure~\ref{fig:advantage_pose}, including both the part maps and the predicted part visibility score. 
       %
       %
       The visualization part maps show that the pose-guided attention mask can focus more on the regions which are not occluded.
       %
       %
       Therefore, we can deduce that, compared with pose cues, the appearance cues can not offer enough insight especially when facing new obstructions.

       \subsection{Parameter Analysis}
         
\begin{figure}
             \centering
             \includegraphics[width=\linewidth]{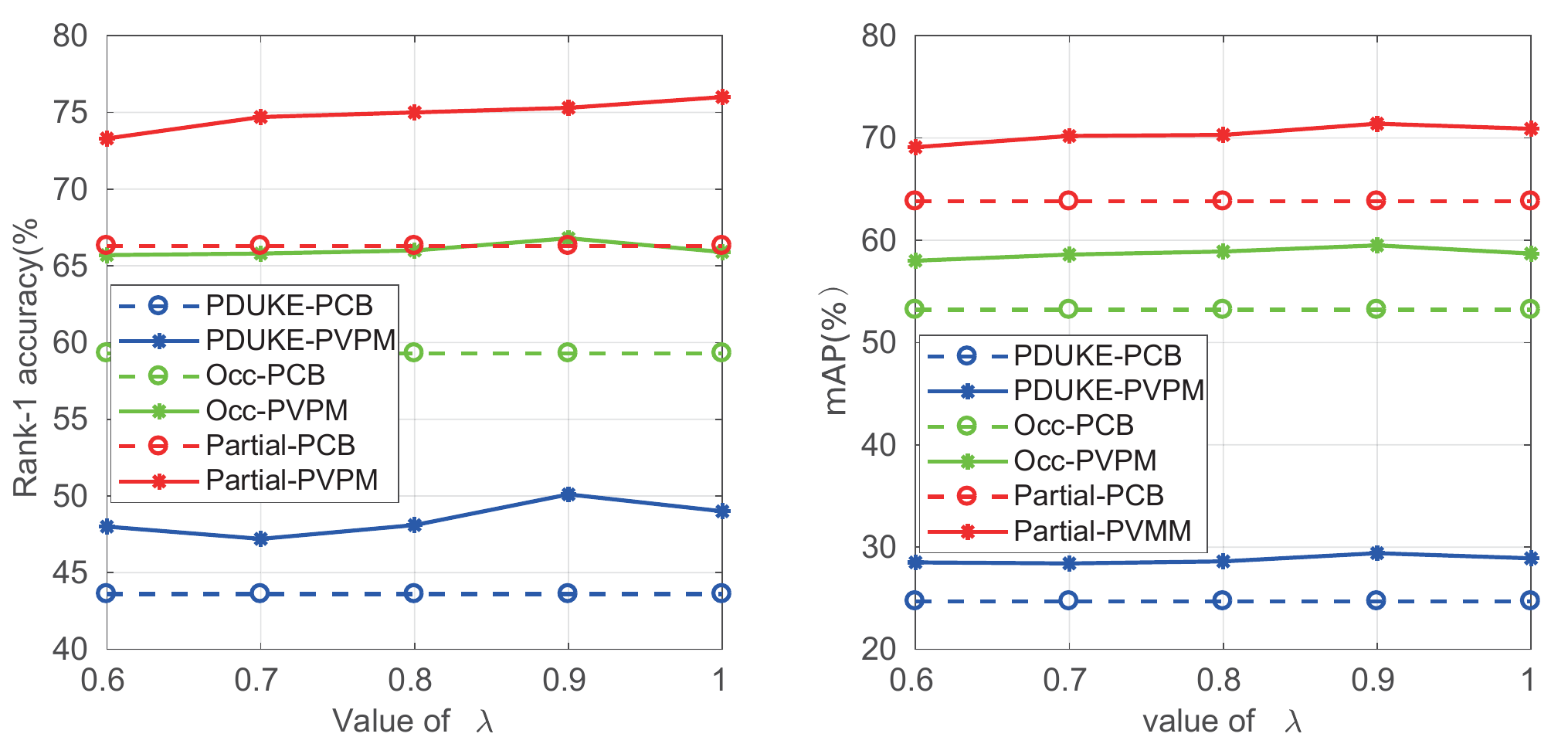}
             \caption{Rank-1 accuracy and mAP with different $\lambda$ settings. The red, green and blue lines correspond to the results of the Partial ReID, Occluded ReID and p-DukeMTMC-reID dataset.}
             \label{fig:lambda}
           \end{figure}

        \noindent\textbf{The Impact of regularization coefficient $\lambda$. }
       $\lambda$ is the regularization coefficient of Eq.\ref{eq:solve_v}. 
       Small $\lambda$ will weaken the discriminative ability of the visibility predictor, leading to all parts thought as visible.
       But a large $\lambda$ may mislead the visibility predictor taking some local regions as unobservable.
       In this section, we compare the performance with different settings of $\lambda$, which varies from 0.6 to 1.
       We show the rank-1 accuracy and mAP variations in Figure.\ref{fig:lambda}.
       As can be seen, the performance reaches the peak value at 0.9, and drop a little bit with $\lambda$ increasing to 1. 
       This performance trend just validates our expectation of the coefficient $\lambda$.
       
\begin{figure}
             \centering
             \includegraphics[width=\linewidth]{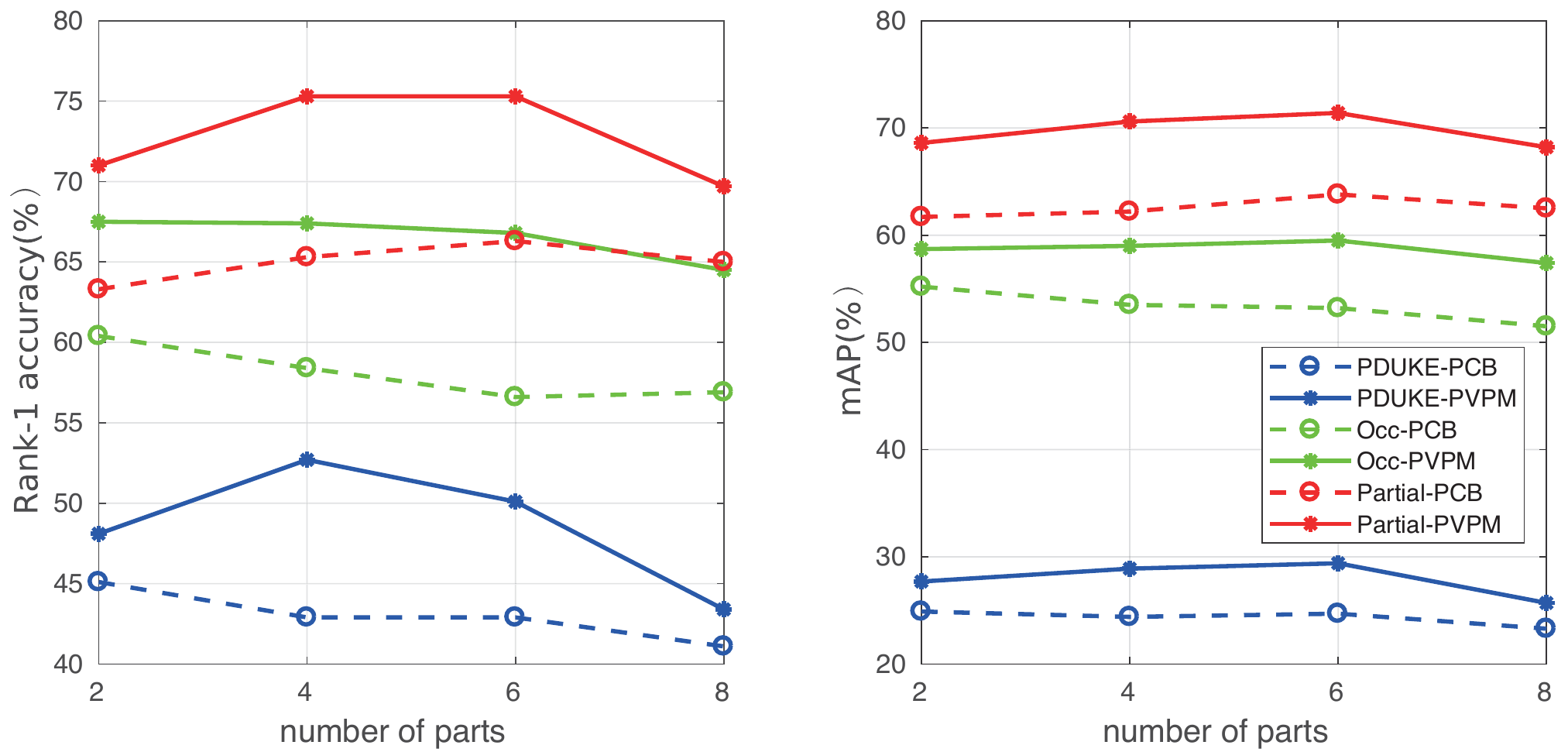}
             \caption{Rank-1 accuracy and mAP comparison with different setting of part number $N_p$.} 
             \label{fig:part_num}
             \vspace{-4mm}
           \end{figure}

       \noindent\textbf{The Impact of Part Number $N_p$.} 
       $N_p$ determines the granularity of the part feature. 
       We conduct several experiments by setting $N_p$ from 2 to 8, and demonstrate the result in Figure~\ref{fig:part_num}, including the rank-1 matching rate and the mAP value. 
       As can be seen, with $N_p$ increases, the performance keeps improves at first, and reaches the peak when $N_p$ arrives 4.
       However, the performance starts to drop with the part number continuing enlarging to 8.
       We suggest that this phenomenon may be drawn by the over-increased $N_p$, making the small parts becoming similar to each other and decreasing the discriminative ability of our model.

      \section{Conclusion}
       In this paper, we propose a novel Pose-guided Visible Part Matching (PVPM) algorithm for occlusion ReID task. The proposed PVPM jointly considers the discriminative pose-guided attention and part visibility in a unified framework.  Unlike most existing methods which utilize visibility cues from other data source directly, we explore the part correspondence on target data and self-mine visibility score via graph matching. A self-learning method was introduced for pseudo label generation and optimize visibility predictor without data bias.
       %
       %
       Sufficient experimental results on the three reported occluded datasets demonstrate the superiority of the proposed model for occluded person ReID task.
      
       \vspace{2mm}
      
       \noindent\textbf{Acknowledgements.} 
       The paper is supported in part by the National Key R\&D Program of China under Grant No.2018AAA0102001 and National Natural Science Foundation of China under Grant No.61725202, U1903215, 61829102, 91538201, 61751212 and the Fundamental Research Funds for the Central Universities under Grant No. DUT19GJ201.

           {\small
           \bibliographystyle{ieee_fullname}
           \bibliography{egbib}

\begin{thebibliography}{10}\itemsep=-1pt

\bibitem{DBLP:conf/cvpr/CaiWC19}
Honglong Cai, Zhiguan Wang, and Jinxing Cheng.
\newblock Multi-scale body-part mask guided attention for person
  re-identification.
\newblock In {\em CVPR Workshops}, 2019.

\bibitem{cao2017realtime}
Zhe Cao, Tomas Simon, Shih-En Wei, and Yaser Sheikh.
\newblock Realtime multi-person 2d pose estimation using part affinity fields.
\newblock In {\em CVPR}, 2017.

\bibitem{Caron_2018_ECCV}
Mathilde Caron, Piotr Bojanowski, Armand Joulin, and Matthijs Douze.
\newblock Deep clustering for unsupervised learning of visual features.
\newblock In {\em ECCV}, 2018.

\bibitem{chang2018multi}
Xiaobin Chang, Timothy~M Hospedales, and Tao Xiang.
\newblock Multi-level factorisation net for person re-identification.
\newblock In {\em CVPR}, 2018.

\bibitem{gray2007evaluating}
Douglas Gray, Shane Brennan, and Hai Tao.
\newblock Evaluating appearance models for recognition, reacquisition, and
  tracking.
\newblock In {\em PETS}, 2007.

\bibitem{he2016deep}
Kaiming He, Xiangyu Zhang, Shaoqing Ren, and Jian Sun.
\newblock Deep residual learning for image recognition.
\newblock In {\em CVPR}, 2016.

\bibitem{he2018recognizing}
Lingxiao He, Zhenan Sun, Yuhao Zhu, and Yunbo Wang.
\newblock Recognizing partial biometric patterns.
\newblock {\em arXiv preprint arXiv:1810.07399}, 2018.

\bibitem{jing2018selfsupervised}
Longlong Jing, Xiaodong Yang, Jingen Liu, and Yingli Tian.
\newblock Self-supervised spatiotemporal feature learning via video rotation
  prediction, 2018.

\bibitem{kalayeh2018human}
Mahdi~M Kalayeh, Emrah Basaran, Muhittin G{\"o}kmen, Mustafa~E Kamasak, and
  Mubarak Shah.
\newblock Human semantic parsing for person re-identification.
\newblock In {\em CVPR}, 2018.

\bibitem{li2018harmonious}
Wei Li, Xiatian Zhu, and Shaogang Gong.
\newblock Harmonious attention network for person re-identification.
\newblock In {\em CVPR}, 2018.

\bibitem{Li_2019_CVPR}
Wei-Hong Li, Fa-Ting Hong, and Wei-Shi Zheng.
\newblock Learning to learn relation for important people detection in still
  images.
\newblock In {\em CVPR}, 2019.

\bibitem{liao2012partial}
Shengcai Liao, Anil~K Jain, and Stan~Z Li.
\newblock Partial face recognition: Alignment-free approach.
\newblock {\em TPAMI}, 2012.

\bibitem{lin2014microsoft}
Tsung-Yi Lin, Michael Maire, Serge Belongie, James Hays, Pietro Perona, Deva
  Ramanan, Piotr Doll{\'a}r, and C~Lawrence Zitnick.
\newblock Microsoft coco: Common objects in context.
\newblock In {\em ECCV}, 2014.

\bibitem{DBLP:conf/cvpr/LiuNYZCH18}
Jinxian Liu, Bingbing Ni, Yichao Yan, Peng Zhou, Shuo Cheng, and Jianguo Hu.
\newblock Pose transferrable person re-identification.
\newblock In {\em CVPR}, 2018.

\bibitem{liu2017hydraplus}
Xihui Liu, Haiyu Zhao, Maoqing Tian, Lu Sheng, Jing Shao, Shuai Yi, Junjie Yan,
  and Xiaogang Wang.
\newblock Hydraplus-net: Attentive deep features for pedestrian analysis.
\newblock In {\em ICCV}, 2017.

\bibitem{miaopose}
Jiaxu Miao, Yu Wu, Ping Liu, Yuhang Ding, and Yi Yang.
\newblock Pose-guided feature alignment for occluded person re-identification.
\newblock 2019.

\bibitem{noroozi2016unsupervised}
Mehdi Noroozi and Paolo Favaro.
\newblock Unsupervised learning of visual representations by solving jigsaw
  puzzles.
\newblock In {\em ECCV}, 2016.

\bibitem{Noroozi_2018_CVPR}
Mehdi Noroozi, Ananth Vinjimoor, Paolo Favaro, and Hamed Pirsiavash.
\newblock Boosting self-supervised learning via knowledge transfer.
\newblock In {\em CVPR}, 2018.

\bibitem{ristani2018features}
Ergys Ristani and Carlo Tomasi.
\newblock Features for multi-target multi-camera tracking and
  re-identification.
\newblock In {\em CVPR}, 2018.

\bibitem{su2017pose}
Chi Su, Jianing Li, Shiliang Zhang, Junliang Xing, Wen Gao, and Qi Tian.
\newblock Pose-driven deep convolutional model for person re-identification.
\newblock In {\em ICCV}, 2017.

\bibitem{suh2015subgraph}
Yumin Suh, Kamil Adamczewski, and Kyoung Mu~Lee.
\newblock Subgraph matching using compactness prior for robust feature
  correspondence.
\newblock In {\em CVPR}, 2015.

\bibitem{suh2018part}
Yumin Suh, Jingdong Wang, Siyu Tang, Tao Mei, and Kyoung Mu~Lee.
\newblock Part-aligned bilinear representations for person re-identification.
\newblock In {\em ECCV}, 2018.

\bibitem{sun2019perceive}
Yifan Sun, Qin Xu, Yali Li, Chi Zhang, Yikang Li, Shengjin Wang, and Jian Sun.
\newblock Perceive where to focus: Learning visibility-aware part-level
  features for partial person re-identification.
\newblock In {\em CVPR}, 2019.

\bibitem{sun2018beyond}
Yifan Sun, Liang Zheng, Yi Yang, Qi Tian, and Shengjin Wang.
\newblock Beyond part models: Person retrieval with refined part pooling.
\newblock In {\em ECCV}, 2018.

\bibitem{Wei_2018_CVPR}
Donglai Wei, Joseph~J. Lim, Andrew Zisserman, and William~T. Freeman.
\newblock Learning and using the arrow of time.
\newblock In {\em CVPR}, 2018.

\bibitem{wojke2017simple}
Nicolai Wojke, Alex Bewley, and Dietrich Paulus.
\newblock Simple online and realtime tracking with a deep association metric.
\newblock In {\em ICIP}, 2017.

\bibitem{zhang2017alignedreid}
Xuan Zhang, Hao Luo, Xing Fan, Weilai Xiang, Yixiao Sun, Qiqi Xiao, Wei Jiang,
  Chi Zhang, and Jian Sun.
\newblock Alignedreid: Surpassing human-level performance in person
  re-identification.
\newblock {\em arXiv preprint arXiv:1711.08184}, 2017.

\bibitem{zhao2017spindle}
Haiyu Zhao, Maoqing Tian, Shuyang Sun, Jing Shao, Junjie Yan, Shuai Yi,
  Xiaogang Wang, and Xiaoou Tang.
\newblock Spindle net: Person re-identification with human body region guided
  feature decomposition and fusion.
\newblock In {\em CVPR}, 2017.

\bibitem{zhao2017deeply}
Liming Zhao, Xi Li, Yueting Zhuang, and Jingdong Wang.
\newblock Deeply-learned part-aligned representations for person
  re-identification.
\newblock In {\em ICCV}, 2017.

\bibitem{zheng2015scalable}
Liang Zheng, Liyue Shen, Lu Tian, Shengjin Wang, Jingdong Wang, and Qi Tian.
\newblock Scalable person re-identification: A benchmark.
\newblock In {\em ICCV}, 2015.

\bibitem{zheng2015partial}
Wei-Shi Zheng, Xiang Li, Tao Xiang, Shengcai Liao, Jianhuang Lai, and Shaogang
  Gong.
\newblock Partial person re-identification.
\newblock In {\em ICCV}, 2015.

\bibitem{zheng2017unlabeled}
Zhedong Zheng, Liang Zheng, and Yi Yang.
\newblock Unlabeled samples generated by gan improve the person
  re-identification baseline in vitro.
\newblock In {\em ICCV}, 2017.

\bibitem{torchreid}
Kaiyang Zhou and Tao Xiang.
\newblock Torchreid: A library for deep learning person re-identification in
  pytorch.
\newblock {\em arXiv preprint arXiv:1910.10093}, 2019.

\bibitem{zhou2019omni}
Kaiyang Zhou, Yongxin Yang, Andrea Cavallaro, and Tao Xiang.
\newblock Omni-scale feature learning for person re-identification.
\newblock {\em arXiv preprint arXiv:1905.00953}, 2019.

\bibitem{DBLP:conf/aaai/ZhouFZSLWL18}
Qin Zhou, Heng Fan, Shibao Zheng, Hang Su, Xinzhe Li, Shuang Wu, and Haibin
  Ling.
\newblock Graph correspondence transfer for person re-identification.
\newblock In {\em AAAI}, 2018.

\bibitem{zhuo2018occluded}
Jiaxuan Zhuo, Zeyu Chen, Jianhuang Lai, and Guangcong Wang.
\newblock Occluded person re-identification.
\newblock In {\em ICME}, 2018.

\bibitem{zhuo2019novel}
Jiaxuan Zhuo, Jianhuang Lai, and Peijia Chen.
\newblock A novel teacher-student learning framework for occluded person
  re-identification.
\newblock {\em arXiv preprint arXiv:1907.03253}, 2019.

\end{thebibliography}
           }
           
           \end{document}